\documentclass{article} 
\usepackage{iclr2020_conference,times}

\usepackage{hyperref}
\usepackage{url}

\usepackage{graphicx}
\usepackage{amsmath}
\usepackage{color}
\usepackage{algorithm}
\usepackage[noend]{algpseudocode}
\usepackage{siunitx}
\usepackage{subcaption}
\usepackage{wrapfig}
\usepackage{listings}
\usepackage[linewidth=1pt]{mdframed}
\usepackage{tikz-dependency}
\usepackage{amsmath, graphicx, amsfonts, amssymb}
\usepackage{algorithmicx}
\usepackage{booktabs}
\usepackage[percent]{overpic}
\usepackage{subcaption}

\usepackage[page]{appendix}

\usepackage{hyperref}
\usepackage{url}
\NewEnviron{elaboration}{
\par
\begin{tikzpicture}
\node[rectangle,minimum width=\textwidth] (m) {\begin{minipage}{.98\textwidth}\BODY\end{minipage}};
\draw[dashed] (m.south west) rectangle (m.north east);
\end{tikzpicture}
}

\title{How To Avoid Being Eaten By a Grue: Exploration Strategies for Text-Adventure Agents}

\author{Prithviraj Ammanabrolu, Ethan Tien, Zhaochen Luo, Mark O. Riedl \\
Georgia Institute of Technology\\
\texttt{\{raj.ammanabrolu, etien, zluo, riedl\}@gatech.edu} \\
}

\iclrfinalcopy 
\begin{document}

\maketitle

\begin{abstract}
Text-based games---in which an agent interacts with the world through textual natural language---present us with the problem of {\em combinatorially-sized action-spaces}.
Most current reinforcement learning algorithms are not capable of effectively handling such a large number of possible actions per turn.
Poor sample efficiency, consequently, results in agents that are unable to pass bottleneck states, where they are unable to proceed because they do not see the right action sequence to pass the bottleneck enough times to be sufficiently reinforced.
Building on prior work using knowledge graphs in reinforcement learning, we introduce two new game state exploration strategies.
We compare our exploration strategies against strong baselines on the classic text-adventure game, {\em Zork1}, where prior agent have been unable to get past a bottleneck where the agent is eaten by a Grue. 
\end{abstract}

\section{Introduction and Background}
\begin{figure*}
    \centering
    \begin{subfigure}{0.45\textwidth}
\begin{mdframed}
\begin{elaboration}
  \parbox{.99\textwidth}{
\emph{Observation:} \textbf{West of House} You are standing in an open field west of a white house, with a boarded front door. There is a small mailbox here.
}
\end{elaboration}
\begin{flushleft}
\emph{Action:} \textbf{Open mailbox}
\end{flushleft}
\begin{elaboration}
  \noindent\parbox{.99\textwidth}{
\emph{Observation:} Opening the small mailbox reveals a leaflet.
}
\end{elaboration}
\begin{flushleft}
\emph{Action:} \textbf{Read leaflet}
\end{flushleft}
\begin{elaboration}
  \noindent\parbox{.99\textwidth}{
\emph{Observation:} (Taken) "WELCOME TO ZORK!
ZORK is a game of adventure, danger, and low cunning. In it you will explore some of the most amazing territory ever seen by mortals. No computer should be without
one!"
}
\end{elaboration}
\end{mdframed}
\caption{Excerpt from the initial stages of {\em Zork1}.}
        \label{fig:zorkexc}

\label{fig:quest}
    \end{subfigure}
    \begin{subfigure}{0.4\textwidth}
    \centering
    \includegraphics[width=0.49\linewidth]{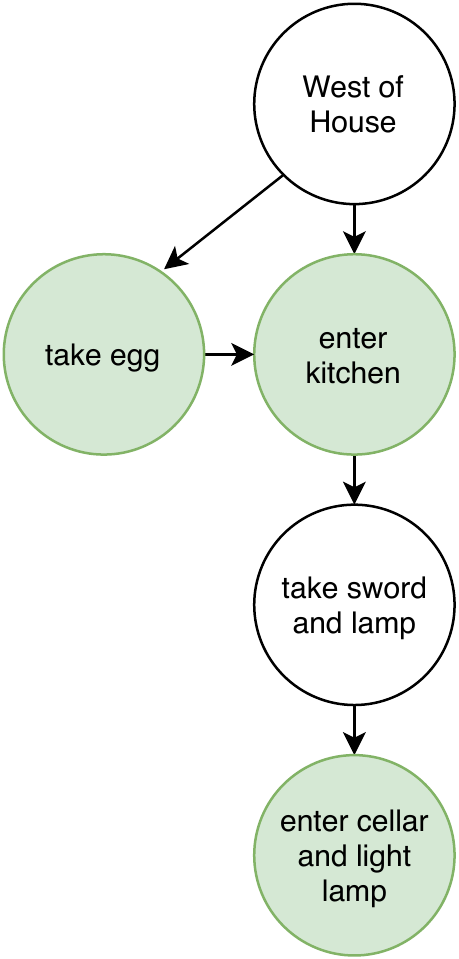}
    \caption{Visualization of the quest structure as a directed acyclic graph in {\em Zork1} demonstrating bottlenecks. Each node represents an action that needs to be taken to finish the quest. Green nodes represent potential positive rewards. By our definition, entering the kitchen and lighting the lamp after entering the cellar are likely bottleneck candidates.}
            \label{fig:dag}

    \end{subfigure}
    \caption{An overall example of an excerpt and quest structure of {\em Zork1}.}
    
\end{figure*}

Many reinforcement learning algorithms are designed for relatively small discrete or continuous action spaces and so have trouble scaling.
Text-adventure games---or interaction fictions---are simulations in which both an agents' state and action spaces are in textual natural language.
An example of a one turn agent interaction in the popular text-game {\em Zork1} can be seen in Fig.~\ref{fig:zorkexc}.
Text-adventure games provide us with multiple challenges in the form of partial observability, commonsense reasoning, and a combinatorially-sized state-action space.
Text-adventure games are structured as long puzzles or quests, interspersed with bottlenecks.
The quests can usually be completed  through multiple branching paths.
However, games can also feature one or more bottlenecks.
Bottlenecks are areas that an agent must pass through in order to progress to the next section of the game regardless of what path the agent has taken to complete that section of the quest~\citep{options}.
In this work, we focus on more effectively exploring this space and surpassing these bottlenecks---building on prior work that focuses on tackling the other problems.

Formally, we use the definition of text-adventure games as seen in~\citet{cote18} and \citet{jericho}.
These games are partially observable Markov decision processes (POMDPs), represented as a 7-tuple of $\langle S,T,A,\Omega , O,R, \gamma\rangle$ representing the set of environment states, mostly deterministic conditional transition probabilities between states, the vocabulary or words used to compose text commands, observations returned by the game, observation conditional probabilities, reward function, and the discount factor respectively.
For our purposes, understanding the exact state and action spaces we use in this work is critical and so we define each of these in relative depth.

\textbf{Action-Space.} 
To solve {\em Zork1}, the cannonical text-adventure games, requires the generation of actions consisting of up to five-words from a relatively modest vocabulary of 697 words recognized by the game’s parser.
This results in $\mathcal{O}(697^5)=\num{1.64e14}$ possible actions at every step.
To facilitate text-adventure game playing, \citet{jericho} introduce Jericho\footnote{\url{https://github.com/microsoft/jericho}}, a framework for interacting with text-games.
They propose a template-based action space in which the agent first selects a template, consisting of an action verb and preposition, and then filling that in with relevant entities $($e.g. $[get]$ \underline{\hspace*{.4cm}}$ [from] $\underline{\hspace*{.4cm}}$)$.
{\em Zork1} has 237 templates, each with up to two blanks, yielding a template-action space of size $\mathcal{O}(237 \times 697^2)=\num{1.15e8}$.
This space is still far larger than most used by previous approaches applying reinforcement learning to text-based games.

\textbf{State-Representation.}
Prior work has shown that knowledge graphs are effective in terms of dealing with the challenges of {\em partial observability}~$($\citeauthor{ammanabrolu}~\citeyear{ammanabrolu};~\citeyear{ammanabrolutransfer}$)$.
A knowledge graph is a set of 3-tuples of the form $\langle subject, relation, object \rangle$.
These triples are extracted from the observations using Stanford's Open Information Extraction (OpenIE)~\citep{Angeli2015}.
Human-made text-adventure games often contain relatively complex semi-structured information that OpenIE is not designed to parse and so they add additional rules to ensure that the correct information is parsed.
The graph itself is more or less a map of the world, with information about objects' affordances and attributes linked to the rooms that they are place in a map.
The graph also makes a distinction with respect to items that are in the agent's possession or in their immediate surrounding environment.
An example of what the knowledge graph looks like and specific implementation details can be found in Appendix~\ref{sec:app-kg}.

\citet{Ammanabrolu2020Graph} introduce the KG-A2C,\footnote{\url{https://github.com/rajammanabrolu/KG-A2C}} which uses a knowledge graph based state-representation to aid in the section of actions in a  combinatorially-sized action-space---specifically they use the knowledge graph to constrain the kinds of entities that can be filled in the blanks in the template action-space.
They test their approach on {\em Zork1}, showing the combination of the knowledge graph and template action selection resulted in improvements over existing methods.
They note that their approach reaches a score of 40 which corresponds to a bottleneck in {\em Zork1} where the player is eaten by a ``grue'' (resulting in negative reward) if the player has not first lit a lamp.
The lamp must be lit many steps after first being encountered, in a different section of the game; this action is necessary to continue exploring but doesn’t immediately produce any positive reward.
That is, there is a long term dependency between actions that is not immediately rewarded, as seen in Figure~\ref{fig:dag}.
Others using artificially constrained action spaces also report an inability to pass through this bottleneck~\citep{zahavy18,jain2019algorithmic}.
They pose a significant challenge for these methods because the agent does not see the correct action sequence to pass the bottleneck enough times.
This is in part due to the fact that for that sequence to be reinforced, the agent needs to reach the next possible reward beyond the bottleneck.


More efficient exploration strategies are required to pass bottlenecks.
Our contributions are two-fold.
We first introduce a method that detects bottlenecks in text-games using the overall reward gained and the knowledge graph state.
This method freezes the policy used to reach the bottleneck and restarts the training from there on out, additionally conducting a backtracking search to ensure that a sub-optimal policy has not been frozen. 
The second contribution explore how to leverage knowledge graphs to improve existing exploration algorithms for dealing with combinatorial action-spaces such as Go-Explore~\citep{ecoffet19}.
We additionally present a comparative ablation study analyzing the performance of these methods on the popular text-game {\em Zork1}.

\section{Exploration Methods}
\label{sec:exploration}
In this section, we describe methods to explore combinatorially sized action spaces such as text-games---focusing especially on methods that can deal with their inherent bottleneck structure.
We first describe our method that explicitly attempts to detect bottlenecks and then describe how an exploration algorithm such as Go Explore~\citep{ecoffet19} can leverage knowledge graphs.

\textbf{KG-A2C-chained}
An example of a bottleneck can be seen in Figure~\ref{fig:dag}.
We extend the KG-A2C algorithm as follows.
First, we detect bottlenecks as states where the agent is unable to progress any further.
We set a {\textit{patience}} parameter and if the agent has not seen a higher score in {\textit{patience}} steps, the agent assumes it has been limited by a bottleneck.
Second, when a bottleneck is found, we freeze the policy that gets the agent to the state with the highest score. 
The agent then begins training a new policy from that particular state.

Simply freezing the policy that led to the bottleneck, however, can potentially result in a policy one that is globally sub-optimal.
We therefore employ a {\em backtracking} strategy that restarts exploration from each of the $n$ previous steps---searching for a more optimal policy that reaches that bottleneck.
%
At each step, we keep track of a buffer of $n$ states and admissible actions that led up to that locally optimal state.
We force the agent to explore from this state to attempt to drive it out of the local optima.
If it is further unable to find itself out of this local optima, we refresh the training process again, but starting at the state immediately before the agent reaches the local optima.
If this continues to fail, we continue to iterate through this buffer of seen states states up to that local optima until we either find a more optimal state or we run out of states to refresh from, in which we terminate the training algorithm.

\textbf{KG-A2C-Explore}
Go-Explore~\citep{ecoffet19} is an algorithm that is designed to keep track of sub-optimal and under-explored states in order to allow the agent to explore upon more optimal states that may be a result of sparse rewards.
The Go-Explore algorithm consists of two phases, the first to continuously explore until a set of promising states and corresponding trajectories are found on the basis of total score, and the second to robustify this found policy against potential stochasticity in the game.
Promising states are defined as those states when explored from will likely result in higher reward trajectories. 
Since the text games we are dealing with are mostly deterministic, with the exception of Zork in later stages, we only focus on using Phase 1 of the Go-Explore algorithm to find an optimal policy.
\cite{madotto2020exploration} look at applying Go-Explore to text-games on a set of simpler games generated using the game generation framework TextWorld~\citep{cote18}.
Instead of training a policy network in parallel to generate actions used for exploration, they use a small set of ``admissible actions''---actions guaranteed to change the world state at any given step during Phase 1---to explore and find high reward trajectories.
This space of actions is relatively small (of the order of $10^2$ per step) and so finding high reward trajectories in larger action-spaces such as in {\em Zork} would be infeasible

Go-Explore maintains an archive of cells---defined as a set of states that map to a single representation---to keep track of promising states.
\cite{ecoffet19} simply encodes each cell by keeping track of the agent's position and \cite{madotto2020exploration} use the textual observations encoded by recurrent neural network as a cell representation.
We improve on this implementation by training the KG-A2C network in parallel, using the snapshot of the knowledge graph in conjunction with the game state to further encode the current state and use this as a cell representation.
At each step, Go-Explore chooses a cell to explore at random (weighted by score to prefer more advanced cells).
The KG-A2C will run for a number of steps, starting with the knowledge graph state and the last seen state of the game from the cell.
This will generate a trajectory for the agent while further training the KG-A2C at each iteration, creating a new representation for the knowledge graph as well as a new game state for the cell.
After expanding a cell, Go-Explore will continue to sample cells by weight to continue expanding its known states.
At the same time, KG-A2C will benefit from the heuristics of selecting preferred cells and be trained on promising states more often.


\section{Evaluation}
We compare our two exploration strategies to the following
baselines and ablations: 

\begin{itemize}
\item
\textbf{KG-A2C} This is the exact same method presented in \cite{Ammanabrolu2020Graph} with no modifications.

\item
\textbf{A2C} Represents the same approach as KG-A2C but with all the knowledge graph components removed.
The state representation is text only encoded using recurrent networks.

\item
\textbf{A2C-chained} Is a variation on KG-A2C-chained where we use our policy chaining approach with the A2C method to train the agent instead of KG-A2C.

\item
\textbf{A2C-Explore} Uses A2C in addition to the exploration strategy seen in KG-A2C-Explore.
The cell representations here are defined in terms of the recurrent network based encoding of the textual observation.
\end{itemize}

\begin{figure*}
\centering
    \begin{subfigure}{0.45\linewidth}
        \centerline{\includegraphics[width=\linewidth]{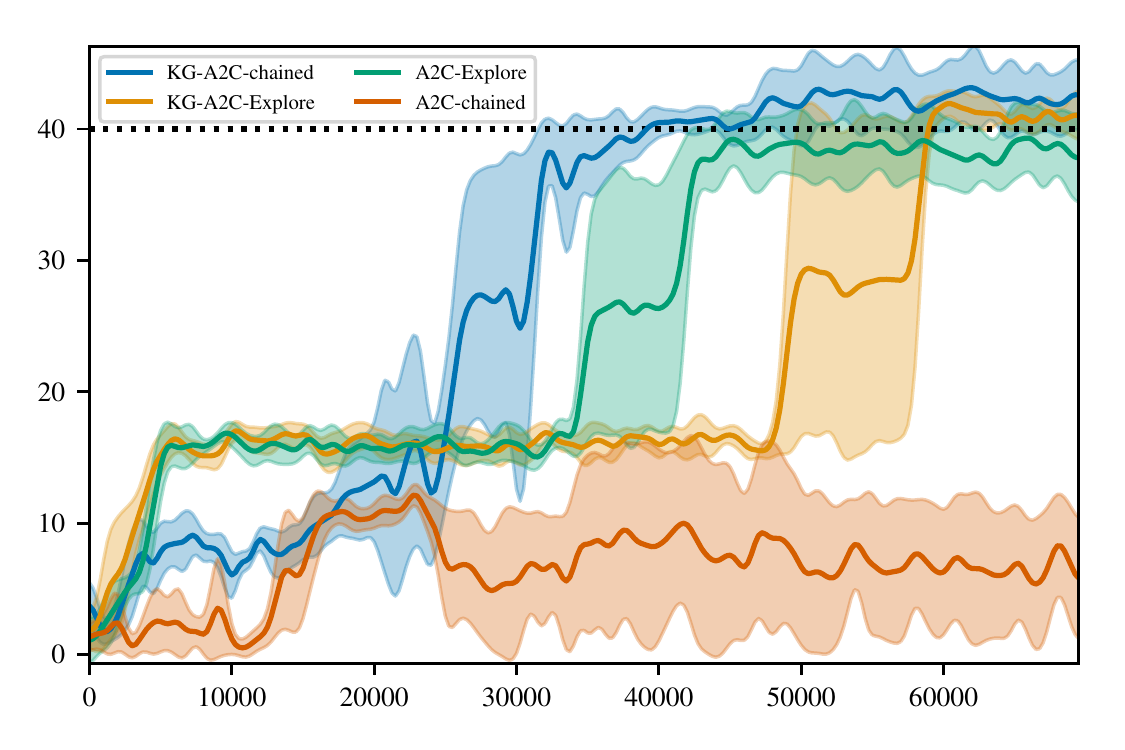}}
        \captionof{figure}{Learning curves for select experiments. The dotted line represents the bottleneck of lighting the lamp.}
        \label{fig:learningcurve}
    \end{subfigure}
    \begin{subfigure}{0.35\linewidth}
        \centering
            \footnotesize

       \begin{tabular}{lc}
         \multicolumn{1}{l}{\textbf{Agent}} & {\textbf{Reward}}  \\ \hline
         A2C                           & 32                                                                                    \\
         KG-A2C                           & 34                                                                                    \\ \hline
         A2C-chained                         & 11.8                                                                                     \\
         KG-A2C-chained                       & 41.8                                                                                    \\ \hline
         A2C-Explore &                40 
                         \\
         KG-A2C-Explore                             & 44
         \end{tabular}
         \caption{Asymptotic reward after training until the exploration algorithms terminate. A2C and KG-A2C are asymptotic rewards reproduced from \cite{Ammanabrolu2020Graph}.}
        \label{tab:zorkrewards}
    \end{subfigure}
\caption{Ablation results on {\em Zork1}, averaged across 5 independent runs.}
\label{fig:zorkresults}
\end{figure*}


Figure~\ref{fig:zorkresults} shows that agents utilizing knowledge-graphs in addition to either enhanced exploration method
far outperform the baseline A2C and KG-A2C. 
KG-A2C-chained and KG-A2C-Explore both pass the bottleneck of a score of 40, whereas A2C-Explore gets to the bottleneck but cannot surpass it.

There are a couple of key insights that can be drawn from these results
%
The first is that the knowledge graph appears to be critical; it is theorized to help with partial observability.
However the knowledge graph representation isn't sufficient in that the knowledge graph representation without enhanced exploration methods cannot surpass the bottleneck.
A2C-chained---which explores without a knowledge graph---fails to even outperform the baseline A2C.
We hypothesize that this is due to the knowledge graph aiding implicitly in the sample efficiency of bottleneck detection and subsequent exploration.
That is, exploring after backtracking from a potentially detected bottleneck is much more efficient in the knowledge graph based agent.

The Go-Explore based exploration algorithm sees less of a difference between agents.
A2C-Explore converges more quickly, but to a lower reward trajectory that fails to pass the bottleneck, whereas KG-A2C-Explore takes longer to reach a similar reward but consistently makes it through the bottleneck.
The knowledge graph cell representation appears to thus be a better indication of what a promising state is as opposed to just the textual observation.

Comparing the advanced exploration methods when using the knowledge graph, we see that both agents successfully pass the bottleneck corresponding to entering the cellar and lighting the lamp and reach comparable scores within a margin of error.
KG-A2C-chained is significantly more sample efficient and converges faster.
We can infer that chaining policies by explicitly detecting bottlenecks lets us pass it more quickly than attempting to find promising cell representations with Go-Explore.
This form of chained exploration with backtracking is particularly suited to sequential decision making problems that can be represented as acyclic directed graphs as in Figure~\ref{fig:dag}.


\bibliography{iclr2020_conference}
\bibliographystyle{iclr2020_conference}

\newpage
\appendix
\section{Appendix}

\subsection{Zork1}

\begin{figure}
    \centering
    \includegraphics[width=\textwidth]{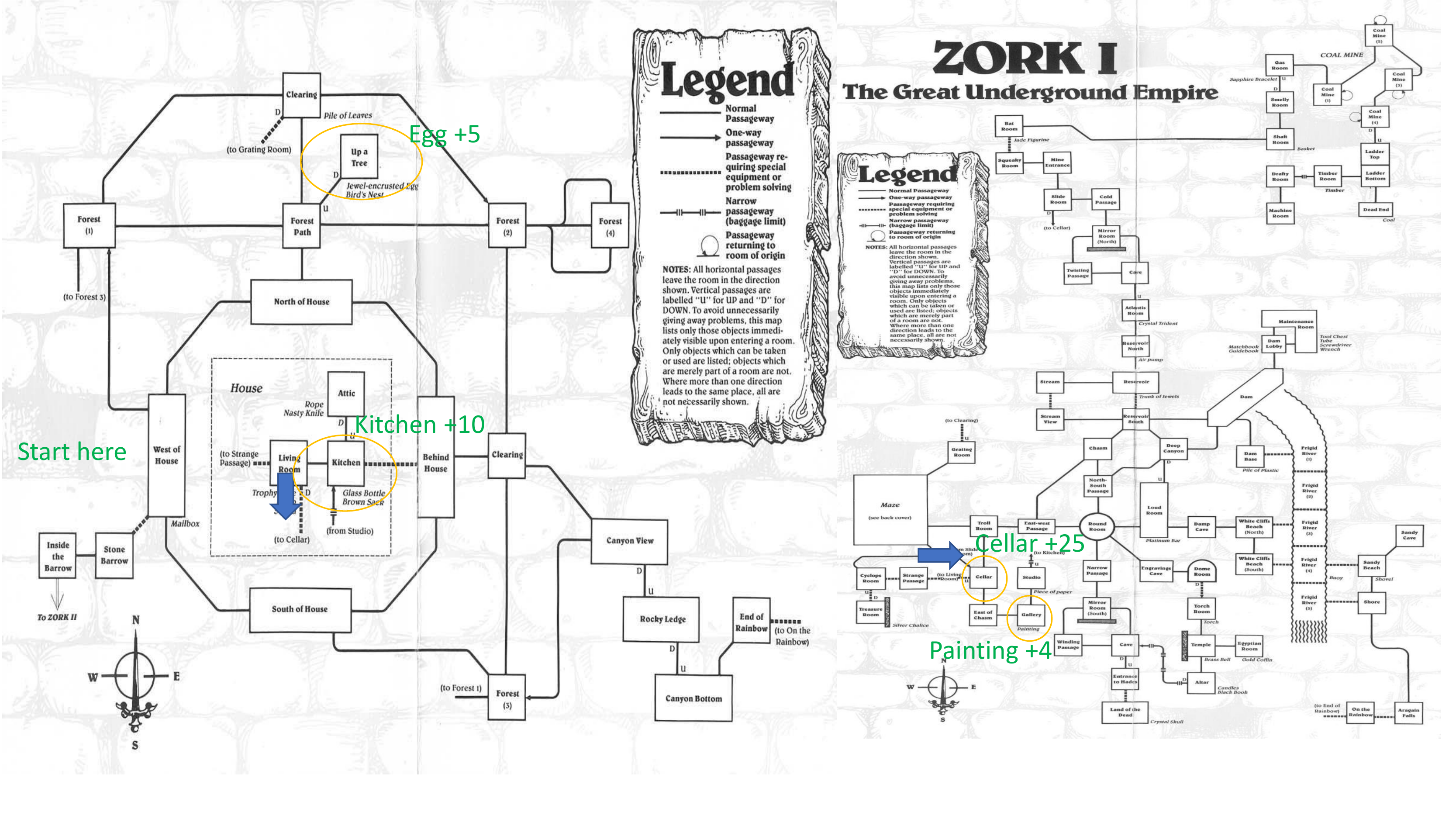}
    \caption{Map of {\em Zork1} annotated with rewards. These rewards correspond to the quest structure seen in Figure~\ref{fig:dag}. Taken from \cite{Ammanabrolu2020Graph}.}
    \label{fig:map}
\end{figure}

{\em Zork1} is one of the first text-adventure games and heavily influences games released later in terms of narrative style and game structure.
It is a dungeon crawler where the player must explore a vast world and collect a series of treasures.
It was identified by~\cite{jericho} as a moonshot game and has been the subject of much work in leaning agents~\citep{jonmay,zahavy18,tessler2019action,jain2019algorithmic}.
Rewards are given to the player when they collect treasures as well as when important intermediate milestones needed to further explore the world are passed.
Figure~\ref{fig:map} and Figure~\ref{fig:dag} show us a map of the world of {\em Zork1} and the corresponding quest structure.

The bottleneck seen at a score of around 40 is when the player first enters the cellar on the right side of the map.
The cellar is dark and you need to immediately light the lamp to see anything.
Attempting to explore the cellar in the dark results in you being instantly killed by a monster known as a ``grue''.

\subsection{Knowledge Graph Rules}
\label{sec:app-kg}
We make no changes from the graph update rules used by~\cite{Ammanabrolu2020Graph}.
Candidate interactive objects are identified by performing part-of-speech tagging on the current observation, identifying singular and proper nouns as well as adjectives, and are then filtered by checking if they can be examined using the command $examine$ $OBJ$.
Only the interactive objects not found in the inventory are linked to the node corresponding to the current room and the inventory items are linked to the ``you'' node.
The only other rule applied uses the navigational actions performed by the agent to infer the relative positions of rooms, e.g. $\langle kitchen,down,cellar \rangle$ when the agent performs $go$ $down$ when in the kitchen to move to the cellar.

\newpage
\subsection{Hyperparameters}
Hyperparameters used for our agents are given below.
{\em Patience} and {\em buffer size} are used for the policy chaining method as described in Section~\ref{sec:exploration}.
{\em Cell step size} is a parameter used for Go-Explore and describes how many steps are taken when exploring in a given cell state.
Base hyperparameters for KG-A2C are taken from \cite{Ammanabrolu2020Graph} and the same parameters are used for A2C.

\centering
\begin{tabular}{l|c}
         \multicolumn{1}{l}{\textbf{Agent}} & {\textbf{Hyperparameters}}  \\ \hline             
         A2C-chained & patience=35\\
          & buffer\_size \textit{n}=40\\
          & batch\_size=32\\
         \hline
         KG-A2C-chained & patience=35\\
          & buffer\_size \textit{n}=40\\ 
          & batch\_size=32\\
         \hline
         A2C-Explore & cell\_step\_size=30\\
          & batch\_size=1\\ 
         \hline
         KG-A2C-Explore & cell\_step\_size=30\\
          & batch\_size=1\\                
         \end{tabular}
        \label{tab:experimenthyperparams}

\end{document}